\pdfoutput=1

\documentclass[11pt]{article}

\usepackage[table,dvipsnames]{xcolor}
\usepackage[final,nopatch=footnote]{microtype}
\usepackage[]{acl}

\usepackage[T1]{fontenc}
\usepackage[utf8]{inputenc}
\usepackage{hyperref}
\usepackage{times}
\usepackage{latexsym}
\usepackage{inconsolata}

\usepackage{amsmath, amssymb}

\usepackage{booktabs}
\usepackage{tabularx}
\usepackage{multirow}
\usepackage{array}
\usepackage{makecell}
\usepackage{siunitx}

\usepackage{graphicx}
\usepackage{caption}
\usepackage{subcaption}

\usepackage{algorithm}
\usepackage{algorithmicx}
\usepackage{algpseudocode}

\usepackage{tcolorbox}

\usepackage{geometry}
\usepackage{setspace}

\title{MTabVQA: Evaluating Multi-Tabular Reasoning of \\ Language Models in Visual Space}

\author{
  \textbf{Anshul Singh\textsuperscript{1}},
  \textbf{Chris Biemann\textsuperscript{2}},
  \textbf{Jan Strich\textsuperscript{2}} \\
  \\
  \textsuperscript{1}Department of Information Technology, Panjab University, India \\
  \textsuperscript{2}Language Technology Group, Universität Hamburg, Germany \\
  \\
  \small{
    \textbf{Correspondence:} \href{mailto:jan.strich@uni-hamburg.de}{jan.strich@uni-hamburg.de}
  }
}

\begin{document}
\maketitle

\begin{abstract}
Vision-Language Models (VLMs) have demonstrated remarkable capabilities in interpreting visual layouts and text. However, a significant challenge remains in their ability to interpret robustly and reason over multi-tabular data presented as images, a common occurrence in real-world scenarios like web pages and digital documents. Existing benchmarks typically address single tables or non-visual data (text/structured). This leaves a critical gap: they don't assess the ability to parse diverse table images, correlate information across them, and perform multi-hop reasoning on the combined visual data. We introduce MTabVQA, a novel benchmark specifically designed for multi-tabular visual question answering to bridge that gap. MTabVQA comprises 3,745 complex question-answer pairs that necessitate multi-hop reasoning across several visually rendered table images. We provide extensive benchmark results for state-of-the-art VLMs on MTabVQA, revealing significant performance limitations. We further investigate post-training techniques to enhance these reasoning abilities and release MTabVQA-Instruct, a large-scale instruction-tuning dataset. Our experiments show that fine-tuning VLMs with MTabVQA-Instruct substantially improves their performance on visual multi-tabular reasoning. Code and dataset\footnote{\href{https://huggingface.co/datasets/mtabvqa/MTabVQA-Eval}{MTabVQA-Eval}} are available online\footnote{\href{https://anonymous.4open.science/r/MTabVQA-EMNLP-B16E}{Anonymous Repository}}.
\end{abstract}

\section{Introduction}
\begin{figure}[!htbp] 
  \centering
  \includegraphics[width=1\linewidth]{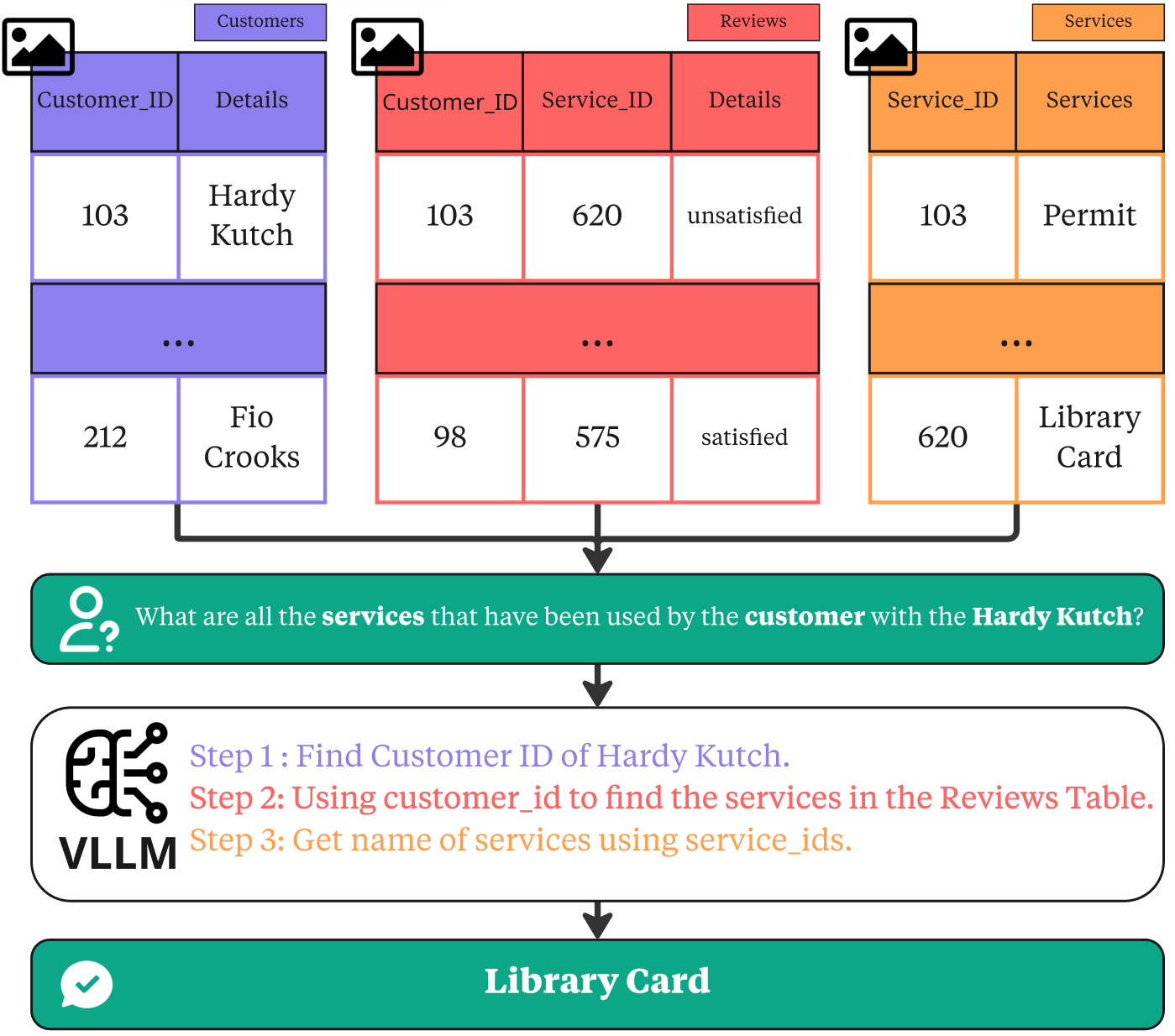} 
  \caption{\small{MTabVQA Benchmark, illustrative example showing three tables (Customers, Reviews, Services), a question requiring multi-table reasoning, the reasoning steps involved, and the final answer derived by a vision-language model.}}
  \label{fig:overview}
\end{figure}

In recent years, vision language models (VLMs) and multimodal systems have demonstrated remarkable capabilities in interpreting complex visual layouts and text \cite{LayoutLLM}, enabling tasks ranging from document understanding \cite{DocKylin}, visual information extraction \cite{GenKIE}, and structured data QA~\cite{AntolALMBZP15} to interactive processes like autonomous web navigation \cite{HeYM0D0L024}. 

Yet, as these models evolve into sophisticated visual agents capable of browsing screen data and performing agentic tasks, a new challenge has emerged: the robust interpretation and reasoning over multi-tabular data presented as images \cite{tableasimages24, multimodaltable24}. This challenge is particularly relevant in real-world scenarios, where tables often appear as images on web pages, PDFs, or digital documents, and extracting actionable insights may require the agent to reference multiple tables simultaneously.

Traditional benchmarks \cite{Spider18, hybridqa20, wikisql17} in table understanding and question answering have primarily focused on single-table scenarios, often relying on textual or HTML representations  \cite{ZhuLHWZLFC20, SuiZZH024}. However, such benchmarks are unable to evaluate model performance on visually complex, multi-tabular data, which requires interpreting layout and structure beyond simple text or HTML. In many practical applications, such as financial analysis, e-commerce, and scientific research \cite{WebTable}, key information is distributed across several tables, each with distinct layouts and visual structures. Current benchmarks \cite{tablebench24, multiTabQA23, wu2025mmqa, MiMoTable}, rooted in single-table, non-visual formats (like text/HTML or relational databases), fail to assess critical capabilities: (1) understanding diverse visual table layouts presented as images, (2) parsing and correlating information across multiple, physically separate tables, and (3) executing multi-hop reasoning grounded in visual data.

To bridge this gap, we propose \textbf{M}ulti-\textbf{Tab}ular \textbf{V}isual \textbf{Q}uestion \textbf{A}nswering (\textbf{MTabVQA}), a novel benchmark specifically designed for assessing the visual reasoning capabilities of models on multi-tabular data represented as images. Distinct from prior benchmarks that primarily focus on single tables \cite{WTQ, wikisql17, multimodaltable24} or utilize non-visual (textual, structured) formats for multi-table reasoning \cite{wu2025mmqa, Spider18, BIRD2023}, MTabVQA uniquely evaluates the integration of information across multiple tables. Our benchmark, comprising \textbf{3,745} question-answer pairs, challenges models with complex queries across \textbf{14 distinct reasoning categories}. These queries are designed to necessitate multi-hop reasoning (e.g., involving aggregation, comparison, or fact-checking) by integrating information from \textbf{two to five} visually table images. MTabVQA enables a targeted evaluation of how well current models handle the process of extracting information from multiple table images and performing the multi-hop reasoning necessary to synthesize answers. Our main contributions are:
\begin{itemize}
    \item We introduce \textbf{MTabVQA}, a novel benchmark designed to evaluate multi-hop reasoning over multiple tables presented as \textbf{images}, addressing a key gap in existing table QA benchmarks.
    \item We provide \textbf{extensive benchmark results} for SOTA open-source and proprietary VLMs on MTabVQA, revealing significant challenges posed by this task.
    \item We release \textbf{MTabVQA-Instruct}, a instruction-tuning dataset. To demonstrate its effectiveness, we introduce \textbf{TableVision}, a VLM fine-tuned on MTabVQA-Instruct, which shows significant improvements on visual multi-tabular reasoning.
\end{itemize}

\begin{table*}[ht]
\centering
\renewcommand{\arraystretch}{1.3}
\resizebox{1\textwidth}{!}{%
\begin{tabular}{l|c|c|c|c|c}
\toprule
\multicolumn{1}{c|}{\textbf{Benchmark}} & \textbf{Question Format} & \textbf{\# Tables/Databases} & \textbf{\# QA Pairs} & \textbf{Task} & \textbf{Modality} \\
\midrule
WTQ \cite{WTQ}                       & NL Questions             & 2,108    & 22,033    & Single-table QA             & Text \\
SQA \cite{sqa17}                       & NL Questions             & N/A      & 17,553    & Single-Table QA             & Text \\
WikiSQL \cite{wikisql17}               & SQL Query                & 24,241   & 80,000+   & Single-table QA             & Text \\ 
Spider \cite{Spider18}                 & NL Questions \& SQL Query & 200      & 10,181    & Text-to-SQL                 & Text \\ 
HybridQA \cite{hybridqa20}               & NL Questions             & 13,000   & 70k       & Table-text QA               & Text \\
FeTaQA \cite{FeTaQA22}                 & NL Questions             & 10,330   & 10k       & Single tables               & Text \\
BIRD \cite{BIRD2023}                   & NL Questions \& SQL Query & 95       & 12,751    & Text-to-SQL                 & Text \\
TableBench \cite{tablebench24}           & NL Questions             & 3,681    & 886       & Single Table                & Text \\ 
SPINACH \cite{spinach24}               & NL Questions \& SQL Query & N/A      & 320       & Text-to-SQL                 & Text \\ 
MMQA \cite{wu2025mmqa}                 & NL Questions \& SQL Query & 3,312    & 3,312     & Text-to-SQL, Multi-table QA & Text \\ 
\midrule 
MMTab \cite{multimodaltable24}           & NL Questions             & 23K      & 49K       & Single-Table QA             & Images \\ %
\rowcolor{blue!20}
\textbf{MTabVQA} (ours)              & NL Questions             & 8499     & 3,745     & Multi-Table QA              & Images \\ %
\bottomrule
\end{tabular}
}
\caption{Differences between our MTabVQA and previous table QA benchmarks. We here abbreviate NL = Natural Language and SQL = Structured Query Language.}
\label{tab:benchmarks_comparision} %
\vspace{-0.2cm}
\end{table*}

\section{Related Work}

Research in table understanding \cite{tablebench24} and multimodal reasoning \cite{multimodaltable24} has advanced significantly. Initial efforts often centered on converting tables into text-based representations like Markdown or HTML \cite{Table-GPT24, TableLlama23}, allowing traditional language models to process them. While effective in controlled environments, this approach encounters limitations in real-world settings where tables frequently appear only as images within documents or web interfaces. Processing visually rendered tables through multi-stage text-conversion pipelines \cite{TableFormer22} presents inherent limitations, they are complex and susceptible to OCR errors, often discard essential visual layout cues (e.g., merged cells, alignment), and risk compounding inaccuracies across stages. This highlights a critical need for models capable of interpreting and reasoning over tables directly from pixel data. 
Moreover, most systems rely on OCR combined with LLMs, which makes them more error-prone compared to developing a single unified model.
Our work focuses squarely on the challenge of extracting information and performing reasoning directly from visual table data, addressing the complexities inherent in image-based table structures.

\subsection{Table Understanding and Extraction}
Effective reasoning over visual tables fundamentally relies on accurate underlying table understanding, including tasks like detection, segmentation, and structure interpretation \cite{BonfittoCM21}. These foundational challenges were often addressed by specialized methods leveraging object detection and OCR, exemplified by systems like TableFormer \cite{TableFormer22}, which improved the extraction of cell structures from images. Despite these advances, such methods frequently encountered difficulties with complex visual layouts and the semantic alignment crucial for interpreting elements like multirow headers or merged cells. Although recent large-scale datasets like MMTab \cite{multimodaltable24} have significantly advanced benchmarking for table extraction and understanding from table images, they primarily focus on single-table scenarios. The challenge of integrating information and reasoning across multiple visually presented tables, which MTabVQA addresses, remains less explored.

\subsection{Multimodal Question Answering}
In parallel, multimodal question answering has made significant progress with models like LLaVA \cite{Llava}, BLIP-2 \cite{Junnan2023}, and GPT-4.1\footnote{\label{gpt41}\href{https://platform.openai.com/docs/models/gpt-4.1}{GPT 4.1}} demonstrating strong capabilities on image-based tasks. While many of these models excel in general visual understanding, they typically treat tabular content as static images, lacking the ability to navigate or reason across multiple tables. Prior benchmarks, such as WikiTableQuestions \cite{WTQ} and WikiSQL \cite{wikisql17}, focus on single-table scenarios and text-based table representations. MMQA \cite{wu2025mmqa}, a recent advancement in this area, extends the evaluation framework to multi-table and multi-hop reasoning. However, MMQA relies on textual inputs rather than raw images.

\subsection{Multi-Tabular Reasoning}
Reasoning across multiple tables demands correlating information from potentially disparate structures via multi-hop operations, a known challenge for current models \cite{multiTabQA23}. While prior work explored multi-table QA \cite{multiTabQA23}, summarization \cite{QFMTS24}, and text-to-SQL \cite{wu2025mmqa}, these efforts predominantly relied on textual or structured data representations. They often bypassed the complexities of interpreting combined visual table layouts, a critical requirement for agents interacting with screen data.
MTabVQA directly addresses this research gap by focusing on \textbf{multi-tabular visual reasoning}. As in Table \ref{tab:benchmarks_comparision}, prominent prior benchmarks like WTQ \cite{WTQ}, WikiSQL \cite{wikisql17}, and even multi-table focused ones such as Spider \cite{Spider18} and MMQA \cite{wu2025mmqa}, primarily operate on textual or structured (e.g., SQL) representations of tables. While MMTab \cite{multimodaltable24} introduced image-based tables, its focus remained on single-table scenarios. In contrast, MTabVQA specifically requires models to answer complex, multi-hop questions by integrating information presented across multiple table images. This necessitates visual parsing of diverse table layouts from images, a capability not comprehensively evaluated by existing benchmarks that are either non-visual or single-table centric. Thus, MTabVQA's unique combination of multi-table reasoning and image-based input directly targets this underexplored area.

\begin{table*}[!ht]
\centering
\begin{resizebox}{0.85\textwidth}{!}{%
  \small 
  \setlength{\tabcolsep}{4pt} 
  \renewcommand{\arraystretch}{1.15}
  \begin{tabular}{@{} l >{\raggedright\arraybackslash}p{3.8cm} l S[table-format=5.0, group-separator={,}] S[table-format=5.0, group-separator={,}] S[table-format=3.1, table-space-text-post=\%] @{}}
  \toprule
  \textbf{Dataset Split} & \textbf{Source} & \textbf{Sub-dataset} & {\textbf{\#QA Pairs}} & {\textbf{\#Tables}} & {\textbf{Proportion (\%)}} \\
  \midrule

  \multirow{5}{*}{\parbox{1.7cm}{\centering\textbf{MTabVQA-Eval}}}
  & QFMTS \citep{QFMTS24}     & MTabVQA-Query  & 2456   & 5541   & 65.7\% \\
  & Spider \citep{Spider18}        & MTabVQA-Spider & 1048   & 2363   & 27.9\% \\
  & Atis \citep{Atis}         & MTabVQA-Atis   &  112   &  429   &  3.0\% \\
  & MiMoTable \citep{MiMoTable} & MTabVQA-Mimo   &  129   &  166   &  3.4\% \\
  \cmidrule(l{2pt}r{2pt}){2-6}
  & \multicolumn{2}{l}{\textbf{Total Eval Set}} & \bfseries \textbf{3745} & \bfseries \textbf{8499} & \bfseries 100.0\% \\
  \midrule

  \multirow{6}{*}{\parbox{1.7cm}{\centering\textbf{MTabVQA-Instruct}}}
  & MultiTabQA \citep{multiTabQA23}            & {--} & 10990  & 21976  & {69.3\%} \\ 
  & Spider \citep{Spider18}   & {--} &  2395  &  5845  & {15.2\% } \\
  & BIRD \cite{BIRD2023}        & {--} &  1572  &  3144  & {9.9\%} \\
  & Atis \cite{Atis}  & {--} &   384  &  1780  & {2.4\%} \\
  & MiMoTable \cite{MiMoTable}         & {--} &   512  &   719  & {3.2\%} \\
  \cmidrule(l{2pt}r{2pt}){2-6} 
  & \multicolumn{2}{l}{\textbf{Full Instruct Set}} & \bfseries \textbf{15853} & \bfseries \textbf{33464} & \bfseries 100.0\% \\
  \bottomrule
  \end{tabular}
}
\caption{Detailed composition of the MTabVQA-Eval and MTabVQA-Instruct datasets. The table shows the original data sources and provides statistics for each sub-dataset, including the number of QA pairs and unique tables.}
\label{tab:mtabvqa_composition}
\vspace{-0.3cm}
\end{resizebox}
\end{table*}

\section{MTabVQA Dataset}
\label{sec:dataset}
We introduce \textbf{M}ulti-\textbf{Tab}ular \textbf{V}isual \textbf{Q}uestion \textbf{A}nswering (\textbf{MTabVQA}), a benchmark specifically designed to assess the capacity of multimodal models to perform multi-hop reasoning across multiple tables presented as images.  MTabVQA dataset includes four sub-datasets based on the primary source databases from which the underlying table data was derived, as detailed in Table \ref{tab:mtabvqa_composition}.
Figure \ref{fig:framework} illustrates the multi-stage process used to construct MTabVQA, encompassing data sourcing, relational analysis, controlled data sampling, image rendering, question-answer pair generation, and rigorous verification.

\subsection{Tabular Data Collection}
\label{sec:source_selection}

MTabVQA utilizes tabular data from diverse sources: BIRD \cite{BIRD2023}, Spider \cite{Spider18}, MiMoTable \cite{MiMoTable}, QFMTS \cite{QFMTS24}, and ATIS \cite{Atis}. We prioritized text-to-SQL datasets as their associated complex SQL queries often involve multi-table joins, naturally lending themselves to multi-table reasoning tasks.

\begin{figure*}[!htbp]
  \centering
  \includegraphics[width=0.95\textwidth]{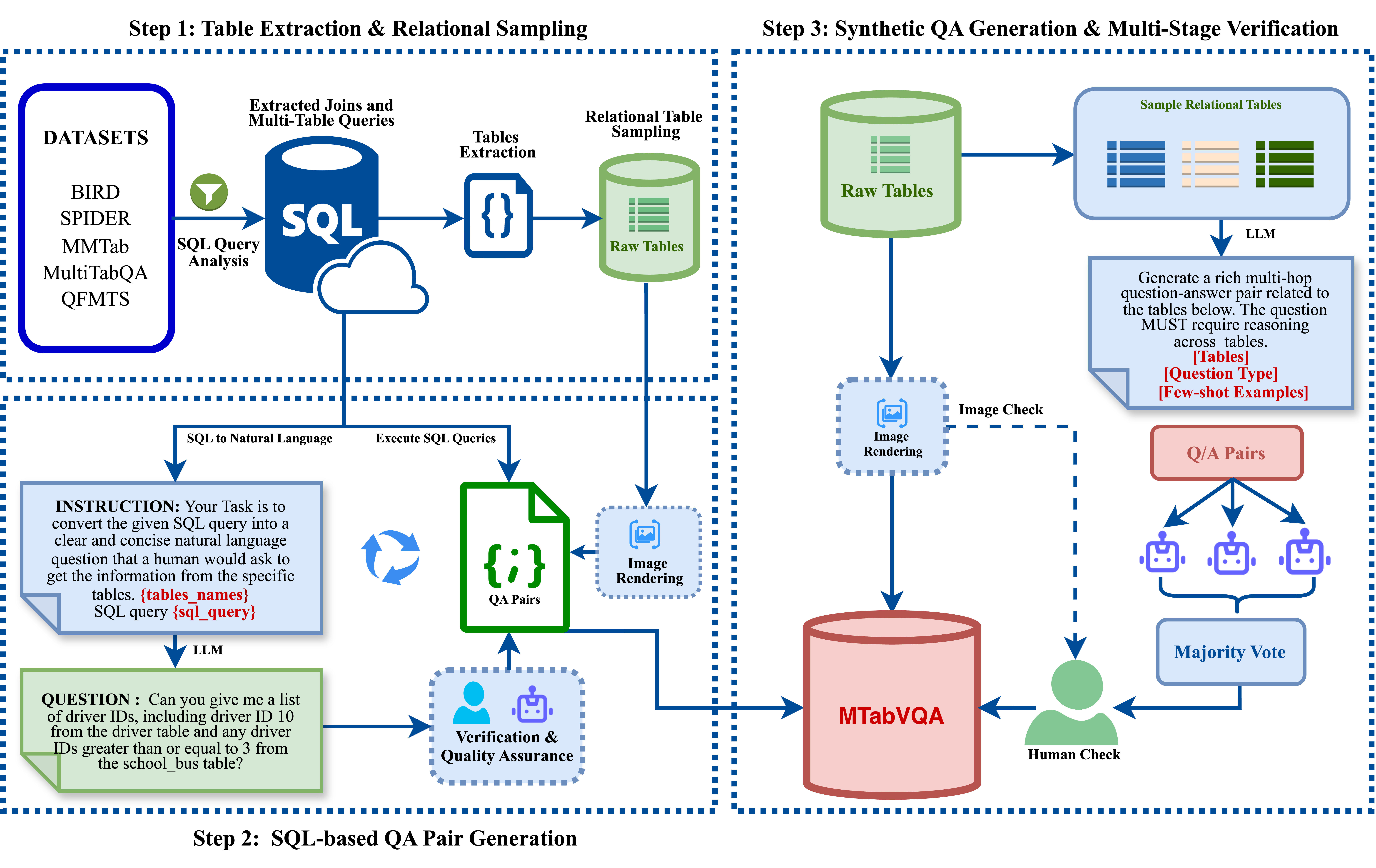}
\caption{\small{MTabVQA Construction Framework Overview. 
  (1) Data Sourcing \& Sampling: Identify multi-table relational data via SQL joins, extract tables, apply relational sampling. 
  (2) Visual QA Generation: Generate multi-hop QA pairs via SQL-to-question conversion or LLM-guided generation from sampled tables/taxonomy; render tables as images. 
  (3) Verification \& Finalization: Apply automated (LLM) and human verification for quality and multi-table necessity.}}
\label{fig:framework}
\end{figure*}

To ensure our benchmark targets multi-table reasoning, we first identified relevant database subsets (Figure \ref{fig:framework}, Step 1). We parsed SQL queries from the source datasets, specifically selecting those requiring multi-table join operations. This analysis confirmed rich inter-table dependencies suitable for our task.
Based on this query analysis, we extracted data instances for the MTabVQA-Eval split: 1,048 multi-join queries from Spider \cite{Spider18} forming MTabVQA-Spider, 2,578 multi-table instances from QFMTS \cite{QFMTS24}, and 112 and 129 multi-table pairs from ATIS \cite{Atis} and MiMoTable \cite{MiMoTable}, respectively. The large and complex BIRD \cite{BIRD2023} dataset, over 7,200 join queries across 69 databases, was primarily used to generate the MTabVQA-Instruct dataset. This query-driven selection ensures that the underlying data inherently necessitates multi-table reasoning.

\vspace{-1em}
\subsection{Data Extraction and Preprocessing}
\label{sec:extraction_preprocessing}

Following the identification of relevant database subsets (Section \ref{sec:source_selection}), we employed a  pipeline to process the data. For each subset, the pipeline extracted the database schemata, including table definitions, column types, primary keys, and foreign key relationships defining inter-table links, and converted the relational data from its native storage (e.g., SQLite) into a standardized JSON format. 
Recognizing that full database tables can be excessively large for visual rendering and efficient model processing, we implemented a controlled sampling strategy. Tables exceeding a predefined row threshold ($N_{max}=50$) were sampled down. While the proportion of excluded data varied depending on the original table sizes in each source dataset, this  threshold aimed to balance visual complexity and data representativeness across the benchmark. 

To preserve crucial relational information between multiple tables during sampling, we utilized a graph-based approach detailed in Algorithm \ref{alg:readable_math_sampling_explained} (Appendix \ref{sec:appendix}). This method ensures referential integrity by preferentially sampling rows linked across related tables via foreign keys, focusing on connections relevant to the multi-table queries identified earlier. The final output for each instance consists of the sampled table data and corresponding schemata, serialized into JSON.

\subsection{Visual Table Rendering}
\label{sec:rendering}

To ensure MTabVQA evaluates visual reasoning over image-based inputs, the sampled tabular data for each QA pair was rendered into images. This step forces models to interpret visual layouts over structured text. We utilized a rendering pipeline employing dataframe\_image\footnote{\href{https://github.com/dexplo/dataframe_image}{dexplo/dataframe\_image}} (with \texttt{selenium} or \texttt{matplotlib} backends) and custom \texttt{Pillow} scripts. This process introduced significant visual diversity by systematically varying structural aspects (e.g., column/row dimensions, relative table positioning) and appearance features (e.g., color schemes, typography, grid styles) across 10 distinct, randomly applied styling themes. This approach simulates the varied appearances of tables in real-world documents and web pages. Further details on the specific themes are provided in Appendix \ref{app:rendering_details}.

\subsection{Multi-Hop QA Pair Generation}
\label{sec:qa_generation}
\begin{figure}[h] 
\centering
\includegraphics[width=0.80\linewidth]{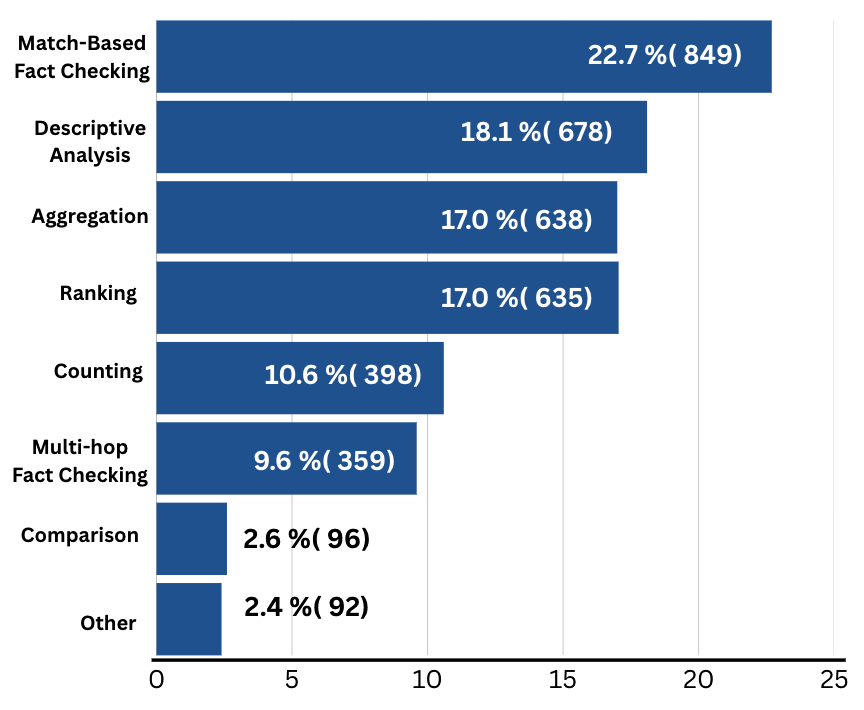} 
\caption{Distribution of Verified Question Categories in the MTabVQA dataset, "Other" includes categories like Anomaly Detection, Arithmetic Calculation, and Multi-hop Numerical Reasoning (total 3,745 QA Pairs).}
\label{fig:composition}
\end{figure}

The pairs of our dataset are designed for multi-hop reasoning across table images, generated via two strategies (Figure \ref{fig:framework}, Steps 2-3): \\
\textbf{1. SQL-to-Question (Step 2):} We converted complex, multi-table SQL queries (from Sec \ref{sec:source_selection}) into natural language questions. For each SQL query, we executed it on sampled table subsets ($S_A, S_B$) for a ground-truth answer. An LLM\footnote{\label{gemini}\href{https://blog.google/technology/google-deepmind/google-gemini-ai-update-december-2024/}{Gemini-2.0-Flash}} then paraphrased the SQL (given schemas and instructions; Figure \ref{fig:framework}, bottom-left prompt) into a question, creating QA pairs grounded in verifiable SQL logic. \\
\textbf{2. Taxonomy-Guided Generation (Step 3):} To diversify reasoning types, an LLM generated novel QA pairs from sampled table subsets and a predefined question taxonomy. This taxonomy, adapted from \cite{tablebench24} to cover common multi-table reasoning patterns (e.g., multi-hop fact-checking, aggregation), guided the LLM (with few-shot examples; Figure \ref{fig:framework}, upper-right prompt) to create questions requiring data from $\ge$2 tables, plus answers and reasoning steps in structured JSON.
Figure~\ref{fig:composition} shows the distribution of the question categories, showing that most of the questions are fact-checking, analysis, aggregation, or ranking.

\subsection{Verification and Filtering}
\label{sec:verification}

\begin{table*}[ht!]
\centering
\resizebox{\textwidth}{!}{
\begin{tabular}{l cccc | cccc | cccc | cccc | cc}
\toprule
\multirow{2}{*}{\textbf{Model}} & \multicolumn{4}{c}{\textbf{MTabVQA-Spider}} & \multicolumn{4}{c}{\textbf{MTabVQA-Query}} & \multicolumn{4}{c}{\textbf{MTabVQA-ATIS}} & \multicolumn{4}{c}{\textbf{MTabVQA-MiMo}} & \multicolumn{2}{c}{\textbf{Overall}} \\
\cmidrule(lr){2-5} \cmidrule(lr){6-9} \cmidrule(lr){10-13} \cmidrule(lr){14-17} \cmidrule(lr){18-19}

 & EM & F1 & P & R & EM & F1 & P & R & EM & F1 & P & R & EM & F1 & P & R & EM & F1 \\
\midrule
\multicolumn{19}{c}{\textit{\textbf{Open-Source VLMs (Zero-Shot)}}} \\
\midrule
Gemma-3-12B-IT & 15.6 & 48.0 & 48.2 & 53.4 & 10.3 & 38.1 & 39.4 & 42.6 & 11.6 & 35.1 & 34.2 & 40.8 & 9.3 & 18.6 & 22.0 & 18.8 & \textbf{11.8} & \textbf{40.1} \\
Qwen2.5-VL-7B & 8.0 & 39.8 & 40.4 & 44.0 & 7.8 & 33.9 & 34.8 & 38.0 & 6.3 & 32.6 & 29.0 & 48.6 & 9.3 & 22.2 & 25.9 & 22.8 & 7.8 & 35.1 \\ 
InternVL3-8B-Instruct & 6.1 & 32.4 & 33.0 & 39.1 & 5.2 & 24.8 & 26.9 & 29.6 & 3.6 & 20.3 & 19.5 & 31.9 & 7.0 & 19.1 & 22.3 & 21.3 & 5.4 & 26.6 \\
Phi-3.5-Vision-Instruct & 2.9 & 26.1 & 25.9 & 39.6 & 2.4 & 22.0 & 22.3 & 34.7 & 1.8 & 15.0 & 15.3 & 24.8 & 0.8 & 3.2 & 3.6 & 3.3 & 2.5 & 22.3 \\ 
LLaVA-OV-Qwen2-7B & 2.2 & 20.0 & 19.5 & 29.3 & 2.3 & 15.7 & 15.9 & 23.6 & 0.0 & 9.2 & 5.9 & 33.8 & 0.7& 5.5& 4.3& 19.1& 2.1& 18.4\\
\midrule
\multicolumn{19}{c}{\textit{\textbf{Proprietary VLMs (Zero-Shot)}}} \\
\midrule
GPT-4.1 & 49.0 & 74.3 & 74.7 & 76.6 & 34.2 & 58.5 & 59.2 & 60.8 & 6.3 & 39.9 & 30.0 & 86.3 & 20.2 & 39.6 & 44.9 & 38.8 & \textbf{37.0} & \textbf{61.7} \\ 
Gemini-2.0-Flash & 42.9 & 68.5 & 69.2 & 71.2 & 31.4 & 57.3 & 58.2 & 60.5 & 22.3 & 36.0 & 37.2 & 37.5 & 24.0 & 42.3 & 49.2 & 41.2 & 34.1 & 59.3 \\ 
\midrule
\multicolumn{19}{c}{\textit{\textbf{Fine-tuned Model (Ours)}}} \\
\midrule
\rowcolor{blue!15}
TableVision (Ours) & 32.4& 64.3 & 66.6& 66.1 & 49.8 & 72.6 & 74.0 & 73.5 & 33.0 & 45.9 & 48.4 & 47.8 & 20.1 & 36.2 & 40.8 & 36.4 & \textbf{43.4} & \textbf{68.2} \\
\bottomrule
\end{tabular}
} 
\caption{Performance Comparison of VLMs on MTabVQA-Eval Splits (\%), and Overall EM/F1 (\%). Models categorized and sorted by overall F1 score within categories. Overall scores are weighted averages. Best overall and best open-source zero-shot overall scores are bolded. EM denotes Exact Match, P Precision, and R Recall.}
\label{tab:benchmarks}
\end{table*}

To ensure QA quality and multi-table focus, our verification process (Figure \ref{fig:framework}, Step 3) was done by automated assessment from three LLM agents\footref{gemini}, guided by a verification prompt (Appendix \ref{app:verification_prompt}). These agents evaluated question validity, multi-hop needs, answer accuracy, reasoning soundness, and multi-table necessity ($\ge$2 tables). LLM outputs (JSON with scores/flags) were aggregated by majority vote.

Pairs meeting criteria (majority valid, confirmed multi-table use, average score $\ge$7.0) advanced to human verification using a Streamlit app (Appendix \ref{fig:streamlit_interface}) for final checks on correctness, especially for complex cases. Human Validation was conducted by one annotator.
Only pairs passing both automated and human checks were integrated into MTabVQA. This LLM-assisted human oversight yielded a high-quality benchmark by filtering invalid tables or incorrect QA pairs.
The resulting verified data formed two entirely disjoint splits, ensuring no overlap between training and evaluation:
\begin{itemize}
    \item \textbf{MTabVQA-Eval:} 3,745 QA pairs for benchmarking VLM performance.
    \item \textbf{MTabVQA-Instruct:} 15,853 instruction-following examples for post-training VLMs.
\end{itemize}

\section{Experiments}
This section details the experiments conducted to evaluate VLM capabilities on visual multi-tabular reasoning using our MTabVQA benchmark. Our experiments encompass three key areas: \\
1.  \textbf{Benchmarking Current VLMs:} We first establish baseline performance by evaluating leading open-source and proprietary VLMs on the MTabVQA-Eval split and compare it with our fine-tuned model. (Section \ref{sec:benchmarking}). \\
2.  \textbf{Evaluating Post-Training Strategies:} We investigate methods to improve VLM performance for multi-table VQA. Using our MTabVQA-Instruct dataset, we explore and compare the effectiveness of different post-training techniques, such as Supervised Fine-Tuning (SFT), Chain-of-Thought (CoT), and Group Relative Policy Optimization (GRPO) \cite{deepseekmath} (Section \ref{sec:post_training_analysis}). \\
3. \textbf{Investigating Impact of Post-training Data Composition:} We further analyze how VLM performance is affected by the scale and source of the data used for instruction fine-tuning (Section \ref{sec:data_scale_source_analysis}). Specifically, we fine-tune models on progressively larger and differently sourced subsets of MTabVQA-Instruct and evaluate their generalization on MTabVQA-Eval (Section \ref{sec:data_scale_source_analysis}).

\subsection{Benchmarking}
\label{sec:benchmarking}
We conducted a comprehensive benchmarking study on MTabVQA-Eval to establish baselines for multi-table visual reasoning. We evaluated leading proprietary VLMs (GPT-4.1\footref{gpt41}, Gemini\footnote{\url{https://aistudio.google.com/}} and prominent open-source alternatives (Qwen2.5 \cite{qwen2.5vl}, Gemma-3 \cite{gemma3}, LLaVA-One-Vision \cite{llava-ov}, InternVL3 \cite{internvl3}, Phi-3.5 \cite{phi3}), alongside our fine-tuned \textbf{TableVision} model. We assessed models in a zero-shot setting across all four MTabVQA-Eval splits (Spider, Query, ATIS, and MiMo), instructing them to generate structured JSON (Appendix \ref{app:standard_eval_prompt}). Generation parameters were set to a temperature of 1.0 and top-P of 1.0.

\textbf{Evaluation Metrics.} We primarily use EM for its strict correctness assessment, especially suitable for factual answers from tables. To capture semantic similarity and partial correctness, we also report F1 score, precision (P), and recall (R), providing a more nuanced view of answer quality.

The results (Table \ref{tab:benchmarks}) highlight MTabVQA's difficulty. Open-source VLMs like LLaVA-One-Vision (2.2\% EM, 16.7\% F1 overall) and Phi-3.5-Vision struggled significantly in zero-shot, with Gemma-3 being the strongest open-source baseline (11.8\% EM, 40.1\% F1 overall). Even proprietary models like GPT-4.1 (37.0\% EM, 61.7\% F1 overall) did not achieve perfect scores and showed performance dips on certain splits (e.g., GPT-4.1 on ATIS scored 6.3\% EM), indicating varied challenges within the benchmark, which shows that there is space for improvement.

\textbf{TableVision}, our model fine-tuned using LoRA (rank 128) on MTabVQA-Instruct with Qwen2.5-VL-7B as its base, demonstrated the value of targeted training by achieving the highest overall performance (43.4\% EM, 68.2\% F1). Notably, TableVision surpassed all other models, including GPT-4.1, on the MTabVQA-Query (49.8\% EM, 72.6\% F1) and MTabVQA-ATIS splits. This shows that  fine-tuning can enable smaller open-source models to outperform larger proprietary systems on complex visual multi-tabular reasoning, underscoring MTabVQA-Instruct's effectiveness.

\subsection{Post-training VLMs for Multi-Table Visual Reasoning}
\label{sec:post_training_analysis}

To explore methods for enhancing VLM performance on visual multi-tabular reasoning, we investigated several post-training techniques using a subset of our MTabVQA-Instruct dataset. Specifically, we utilized 2,395 QA pairs derived from the Spider data source, selected for its demonstrated fine-tuning effectiveness (Section \ref{sec:data_scale_source_analysis}) and manageable size for these intensive experiments. We selected the Qwen2.5-VL-3B model \cite{qwen2.5vl} as our base VLM, primarily due to the significant computational requirements associated with advanced post-training methods like reinforcement learning. Our investigation compared the effectiveness of different post-training techniques for multi-tabular visual reasoning. All evaluations were conducted on a corresponding MTabVQA-Eval split.

\begin{figure}[!t]
  \centering
  \includegraphics[width=1\linewidth]{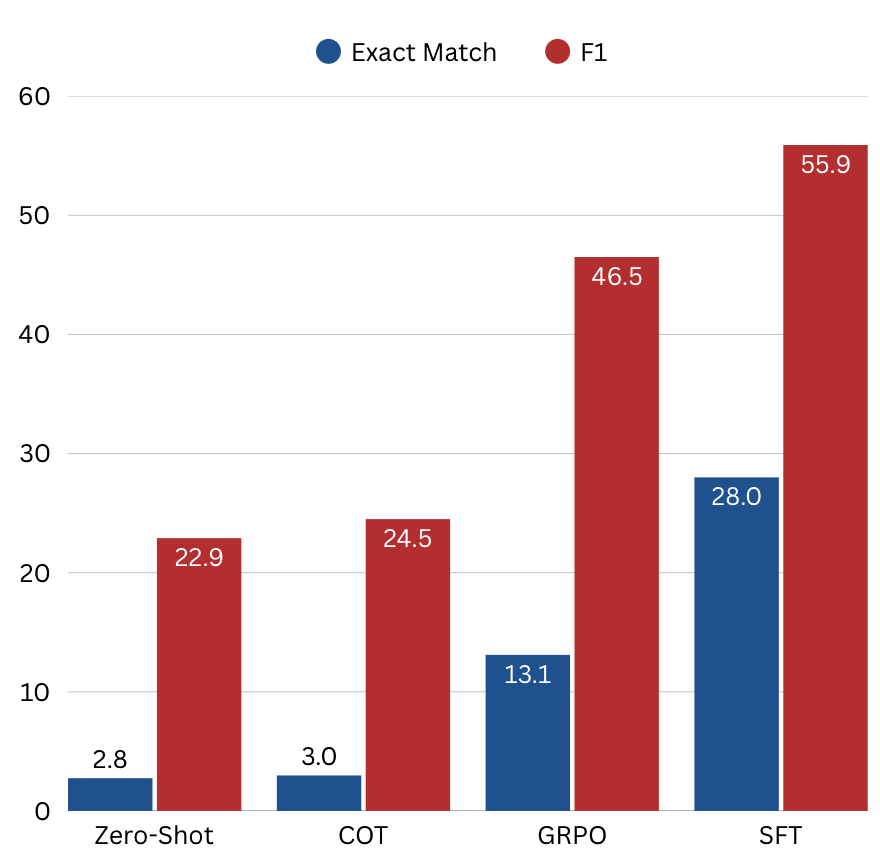}
  \caption{Performance comparison of Qwen2.5-VL-3B on the MTabVQA-Eval using different post-training strategies.}
  \label{fig:post_training_results} %
\end{figure}

First, we established a baseline by evaluating the zero-shot performance of the 3B model. Consistent with observations for larger models (Section \ref{sec:benchmarking}), the base 3B model exhibited poor initial performance on this complex multi-hop reasoning task, achieving an EM of 2.8\% and an F1 score of 22.9\% (Figure \ref{fig:post_training_results}). We then evaluated the efficacy of  using step-by-step reasoning through Chain-of-Thought (CoT) prompting (See Appendix \ref{app:cot_eval_prompt}). While this approach encouraged structured responses, it resulted in only marginal improvements, with EM increasing slightly to 3.0\% and F1 to 24.5\%.

Next, recognizing the reasoning-intensive nature of multi-tabular VQA, we investigated GRPO \cite{deepseekmath}, a reinforcement learning-based post-training approach using the selected 2,395-pair MTabVQA-Instruct subset. As shown in Figure \ref{fig:post_training_results}, GRPO improved performance over the CoT baseline, achieving an EM of 13.1\% and an F1 score of 46.5\%.

Subsequently, we performed SFT on the same subset. For this, we employed LoRA \cite{LoRA} with a rank of 128 for parameter-efficient optimization. SFT yielded substantial performance gains over both CoT and GRPO, boosting EM to 28.0\% and F1 to 55.9\% (Figure \ref{fig:post_training_results}). This demonstrates the strong effectiveness of targeted instruction tuning with SFT for this task in our experiments. While GRPO showed improvement, its gains did not surpass SFT with LoRA. We hypothesize that the effectiveness of GRPO in this context might be limited by the challenge of defining a more sophisticated reward function than a simple exact match/F1 score, which could better capture nuanced aspects of visual multi-tabular reasoning.

\begin{table*}[ht!]
\centering
\resizebox{1\textwidth}{!}{%
\begin{tabular}{l c | cc | cc | cc | cc | cc}
\toprule
\multirow{2}{*}{\textbf{Fine-tuning Subset (Source)}} & \multirow{2}{*}{\textbf{\# Samples}} & \multicolumn{2}{c}{\textbf{MTabVQA-Spider}} & \multicolumn{2}{c}{\textbf{MTabVQA-Query}} & \multicolumn{2}{c}{\textbf{MTabVQA-ATIS}} & \multicolumn{2}{c}{\textbf{MTabVQA-MiMo}} & \multicolumn{2}{c}{\textbf{Overall}} \\
\cmidrule(lr){3-4} \cmidrule(lr){5-6} \cmidrule(lr){7-8} \cmidrule(lr){9-10} \cmidrule(lr){11-12}
 &  & EM & F1 & EM & F1 & EM & F1 & EM & F1 & EM & F1 \\
\midrule
Qwen2.5-VL-7B (Zero-Shot) & 0 & 8.0 & 39.8 & 7.8 & 33.9 & 6.3 & 32.6 & 9.3 & 22.2 & 7.8 & 35.1 \\
\midrule
MiMo+ATIS Subset & 896 & 13.7 & 45.7 & 11.5 & 37.5 & 35.7 & \underline{46.5} & 17.1 & \underline{39.7} & 13.0 & 40.0 \\
Spider Subset & 2,395 & 26.9 & 59.2 & 49.8 & 71.2 & 13.4 & 22.5 & 17.1 & 31.9 & 41.5 & 65.2 \\
MultiTabQA Subset & 10,990 & 10.1 & 33.2 & 8.7 & 28.6 & 16.1 & 41.9 & 11.6 & 25.5 & 9.4 & 30.2 \\
\textbf{MTabVQA-Instruct (Full)}  & 15,853 & 32.4 & \underline{64.3} & 49.8 & \underline{72.6} & 33.0 & 45.9 & 20.2 & 36.2 & \textbf{43.4} & \textbf{68.2} \\
\bottomrule
\end{tabular}%
}
\caption{Performance of fine-tuned models on dataset splits of MTabVQA-Instruct measuring the influence of dataset on the overall performance on MTabVQA-Eval. Performance is measured in EM and F1. \textbf{Bold} indicates the best overall performance. \underline{Underline} indicates best performance for each MTabVQA-Eval subset.}
\label{tab:data_scale_results}
\vspace{-0.2cm}
\end{table*}

\subsection{Impact of Post-training Data Scale and Source}
\label{sec:data_scale_source_analysis}

To understand how instruction-tuning data composition affects performance, we used Qwen2.5-VL-7B as our base VLM. We then fine-tuned it on several MTabVQA-Instruct subsets, each derived from different original data sources (Table \ref{tab:mtabvqa_composition}) to vary both data scale and origin. These models were fine-tuned using Supervised Fine-Tuning (SFT) with LoRA (rank 128) and benchmarked on the full MTabVQA-Eval suite. Table \ref{tab:data_scale_results} presents the EM and F1 scores across MTabVQA-Eval's sub-splits (Spider, Query, ATIS, MiMo) and overall. The fine-tuning subsets included a combined MiMo+ATIS set (896 examples), a Spider-derived set (2,395 examples), a MultiTabQA-derived set (10,990 examples), and our full MTabVQA-Instruct (15,853 examples).

The fine-tuning experiments, detailed in Table \ref{tab:data_scale_results}, reveal a complex relationship between data scale, source, and model performance. Generally, more fine-tuning data leads to better EM and F1 scores, as seen when comparing the MiMo+ATIS subset (896 examples) to the larger Spider subset (2,395 examples). The model trained on the full MTabVQA-Instruct dataset of 15,853 diverse examples achieved the highest overall F1 score (68.2\%), highlighting the benefit of scale when combined with relevant and varied data.

However, the source of the fine-tuning data is critically important. The model trained only on the large MultiTabQA subset exhibited surprisingly low overall performance (30.2\% F1), significantly underperforming compared to the model trained on the much smaller Spider subset and even the MiMo+ATIS subset. This suggests that the characteristics of the MultiTabQA data, while extensive, may not align well with the broader MTabVQA-Eval benchmark or could introduce a domain shift. For instance, its F1 score on MTabVQA-Query and MTabVQA-Spider was substantially lower than that achieved by TableVision or the Spider-tuned model. This highlights that a large volume of data from a single, potentially narrowly focused or misaligned source can be less effective than smaller, more targeted, or diverse datasets.

Furthermore, domain-specific alignment proves beneficial. The model fine-tuned on the Spider subset, for example, demonstrated strong performance on the MTabVQA-Spider eval split. The superior overall performance of TableVision, trained on full MTabVQA-Instruct, indicates that data diversity is crucial for generalization across varied multi-table reasoning scenarios. This shows that while scaling instruction data is generally advantageous, the relevance and diversity of this data with the target tasks is important for achieving optimal performance.

\section{Conclusion}
In this work, we introduce MTabVQA-Eval, a novel and challenging benchmark specifically designed to evaluate the multi-tabular reasoning capabilities of vision-language models over tables presented as images. MTabVQA-Eval, comprising 3,745  QA pairs, focuses on a critical yet underexplored area of integrating and reasoning about information distributed across several table images. This benchmark significantly contributes to bridging the gap between existing table QA benchmarks, which often rely on single or non-visual tables.
We evaluated a range of SOTA open-source and proprietary VLMs on MTabVQA-Eval, revealing substantial challenges these models face with visual multi-tabular reasoning. To address this, we also release MTabVQA-Instruct, a large-scale instruction-tuning dataset. Our experiments demonstrate that our fine-tuned model, TableVision on the MTabVQA-Instruct dataset, leads to considerable performance improvements on this task. Despite these advancements, the performance of VLMs on MTabVQA-Eval indicates significant room for growth, underscoring the complexities of robust visual multi-tabular reasoning and highlighting key areas for future research in developing more capable VLMs.

In future work, we plan to explore more programmatically generated or real-world sourced table images exhibiting even greater visual diversity and degradation to more rigorously test VLM visual parsing and grounding capabilities.

\newpage
\section*{Limitations}
While MTabVQA represents a significant step towards evaluating visual multi-tabular reasoning, we acknowledge several limitations.

\paragraph{English-Only.}
The current iteration of MTabVQA is primarily English-centric. Its underlying tabular data, generated questions, and answers are predominantly in English, which limits the benchmark's applicability for evaluating VLMs on multi-tabular reasoning in other languages. Extending MTabVQA to include multilingual tables and queries would be a valuable contribution, allowing for a more comprehensive assessment of VLM capabilities across diverse linguistic contexts and promoting research in multilingual visual document understanding.

\paragraph{Synthetic Table Layout.}
While MTabVQA tasks require multi-hop reasoning across table images and incorporate varied visual renderings, the scope of this visual complexity could be further expanded. Real-world documents often contain tables with highly unconventional layouts, extensive cell merging/spanning, embedded charts or icons within cells, and varying image quality (e.g., scanned documents with noise), which makes the task even more challenging for LLMs.

\paragraph{Limited Annoation.}
To verify that the QA pairs were correct, we used only one annotator to verify the judgments of the LLM's agent. Although the annotation was carried out carefully, there may have been minimal errors in the data annotation, as there was no double-checking by two people.

\bibliography{custom}

\clearpage 
\onecolumn 
\appendix
\section{Relational Table Sampling}
\label{sec:appendix}
Algorithm \ref{alg:readable_math_sampling_explained} details our method for creating smaller, interconnected samples from large databases.
We start by randomly selecting a limited number of rows (up to a maximum, $N_{max} = 50$) from one initial table. Then, using the database's foreign keys, we identify other tables linked to this first one. When sampling from these linked tables, the crucial step is to find and prioritize rows that are directly related to the rows already chosen from the previous table. This is achieved by matching values in the specific columns that link the tables. This process of finding related data and sampling continues as the algorithm explores outwards to other connected tables, ensuring the final set of sampled tables forms a related subset of the original database.
\vspace{5pt}
\begin{algorithm*}[!h]
\caption{Relational Table Sampling}
\label{alg:readable_math_sampling_explained}
\begin{algorithmic}[1]

\Statex \textbf{Input:}
\Statex \quad $\mathcal{D}$: Input database (collection of tables)
\Statex \quad $\mathcal{R}$: Set of foreign key relationships between tables in $\mathcal{D}$
\Statex \quad $N_{max}$: Maximum number of rows per sampled table
\Statex \quad ($V$: Set of table identifiers derived from $\mathcal{D}$)
\Statex \quad ($G=(V,E)$: Relationship graph derived from $\mathcal{D}$ and $\mathcal{R}$)
\Statex \textbf{Output:}
\Statex \quad $\mathcal{S}$: Set of pairs $(t, S_t)$, where $S_t$ is the sampled row subset for table $t \in V$
\State $\mathcal{S} \leftarrow \emptyset$; $\mathcal{P} \leftarrow \emptyset$ \Comment{$\mathcal{S}$: Output samples, $\mathcal{P}$: Processed tables set}
\State $t_{start} \leftarrow \text{SelectSeed}(V, G)$ \Comment{Select a starting table (e.g., highest degree)}
\State $S_{t_{start}} \leftarrow \text{Sample}(t_{start}, N_{max})$ \Comment{Sample initial rows for $t_{start}$}
\State $\mathcal{S} \leftarrow \{(t_{start}, S_{t_{start}})\}$; $\mathcal{P} \leftarrow \{t_{start}\}$ \Comment{Update output set and processed set}
\State Initialize $Q$; $Q.\text{Enqueue}(t_{start})$ \Comment{$Q$: Queue for Breadth-First Search (BFS)}

\While{$Q$ is not empty} \Comment{Perform BFS traversal}
    \State $t_{curr} \leftarrow Q.\text{Dequeue}()$ \Comment{$t_{curr}$: Current table being processed}
    \For{each $t_{rel} \in \text{Neighbors}(t_{curr}, G) \setminus \mathcal{P}$} \Comment{$t_{rel}$: Related, unprocessed neighbor table}
        \State $R_{linked} \leftarrow \text{GetLinkedRows}(t_{rel}, t_{curr}, S_{t_{curr}}, \mathcal{R})$ \Comment{Get rows in $t_{rel}$ linked to sampled rows}
        \State $S_{t_{rel}} \leftarrow \text{SampleSubset}(R_{linked}, N_{max})$ \Comment{Sample a subset from linked rows, max size $N_{max}$}
        \State $\mathcal{S} \leftarrow \mathcal{S} \cup \{(t_{rel}, S_{t_{rel}})\}$ \Comment{Add the new sample to the output}
        \State $\mathcal{P} \leftarrow \mathcal{P} \cup \{t_{rel}\}$; $Q.\text{Enqueue}(t_{rel})$ \Comment{Mark $t_{rel}$ as processed and add to queue}
    \EndFor
\EndWhile
\State \Return $\mathcal{S}$ \Comment{Return the final set of sampled table subsets}
\end{algorithmic}
\end{algorithm*}

\clearpage
\section{Data Sourcing: Join and Filter Details}
\label{app:data_sourcing_joins}

This section provides a detailed breakdown of the process used to identify and filter data instances requiring multi-table join operations from the source datasets, as mentioned in Section \ref{sec:source_selection}. This formed the basis for constructing both the MTabVQA-Eval and MTabVQA-Instruct splits, ensuring a focus on multi-tabular reasoning. The primary method involved parsing SQL queries associated with text-to-SQL datasets to detect explicit join clauses (e.g., `JOIN`, `INNER JOIN`, `LEFT JOIN`). For datasets without explicit SQL, we relied on provided metadata or question characteristics indicative of multi-table requirements.

\subsection{Spider Dataset}
The Spider dataset \cite{Spider18} is a large-scale text-to-SQL benchmark. We analyzed its train, development (dev), and test splits to identify questions whose corresponding SQL queries involved joins.
\begin{itemize}
    \item \textbf{Train Split:}
        \begin{itemize}
            \item Total Questions: 7,000
            \item Questions with SQL Joins: 2,771
            \item Selected for MTabVQA-Instruct (after filtering and processing): 2,395 instances. 
        \end{itemize}
    \item \textbf{Development (Dev) Split:}
        \begin{itemize}
            \item Total Questions: 1,034
            \item Questions with SQL Joins: 408
        \end{itemize}
    \item \textbf{Test Split:}
        \begin{itemize}
            \item Total Questions: 2,147
            \item Questions with SQL Joins: 862
        \end{itemize}
    \item \textbf{MTabVQA-Eval (from Spider Dev/Test):}
        \begin{itemize}
            \item Combined Join Questions from Dev \& Test: 408 (Dev) + 862 (Test) = 1,270
            \item Selected for MTabVQA-Eval (MTabVQA-Spider-Eval split): 1,048 instances. These were chosen from the 1,270 join questions based on criteria ensuring clear multi-hop reasoning paths, unambiguous answers from sampled data, and visual representability.
        \end{itemize}
\end{itemize}

\subsection{QFMTS Dataset}
The QFMTS dataset \cite{QFMTS24} focuses on query-focused multi-document summarization with tables. We identified instances requiring information synthesis across multiple tables.
\begin{itemize}
    \item Total Questions/Instances: 4,908
    \item Instances Identified as Requiring Multi-Table Reasoning (e.g., via SQL joins or inherent task nature): 2,578
    \item Selected for MTabVQA-Eval (MTabVQA-Query-Eval split, primarily from QFMTS): 2,456 instances. Filtering ensured complexity and suitability for our visual QA benchmark.
\end{itemize}

\subsection{BIRD Dataset}
BIRD \cite{BIRD2023} is another challenging text-to-SQL benchmark designed to evaluate robustness on large databases and complex queries.
\begin{itemize}
    \item Total Identified SQL Join Queries (approx.): 7,900
    \item Generated QA pairs for MTabVQA-Instruct: 1,572 instances. These were generated from a diverse selection of the join queries, focusing on creating complex multi-hop reasoning scenarios suitable for instruction tuning.
\end{itemize}

\subsection{MultiTabQA Dataset}
The MultiTabQA dataset \cite{multiTabQA23} is specifically designed for question answering over multiple tables.
\begin{itemize}
    \item Total QA pairs involving joins/multi-table lookups utilized: 10,990
    \item These were directly incorporated into the MTabVQA-Instruct dataset due to their inherent multi-table nature.
\end{itemize}

\subsection{ATIS Dataset}
The Air Travel Information System (ATIS) dataset \cite{Atis} contains spoken language queries related to flight information, often mapped to relational database queries.
\begin{itemize}
    \item Total Questions Analyzed: 496 
    \item Instances identified/selected for MTabVQA-Eval (MTabVQA-Atis split): 112
    \item Instances selected/generated for MTabVQA-Instruct: 384 (See Table 2).
\end{itemize}

\subsection{MiMoTable Dataset}
The MiMoTable dataset \cite{MiMoTable} focuses on multimodal table understanding.
\begin{itemize}
    \item Total Questions/Instances: 1,636
    \item Questions Identified with Multi-Table Requirements (e.g., from problem descriptions or metadata indicating cross-table information needed): 641
    \item Selected for MTabVQA-Instruct: 512 instances.
    \item Selected for MTabVQA-Eval: 129 instances.
\end{itemize}

\subsection{Overall Summary}
Across all source datasets, we identified approximately \textbf{26,826} potential questions or instances that involved multi-table join operations or inherently required multi-table reasoning. Through our processing, filtering, and generation pipeline, a total of \textbf{19,608} high-quality, multi-tabular visual question-answering instances were curated to form the MTabVQA-Eval (3,745 pairs) and MTabVQA-Instruct (15,853 pairs, with some overlap in underlying source tables but disjoint QA pairs) datasets. The filtering criteria included ensuring genuine multi-hop reasoning, clarity of questions and answers, visual representability of the involved tables, and overall quality for benchmarking and instruction tuning.

\clearpage
\section{Verification Prompt}
\label{app:verification_prompt}

The following prompt was provided to the verification LLMs-based verification agents during the automated assessment phase described in Section \ref{sec:verification}.
\begin{figure}[!htbp] 
\centering
\begin{tcolorbox}[
    width=0.95\linewidth,
    colback=gray!5!white, 
    boxrule=0.5pt,         %
    arc=1mm,               %
    fontupper=\small,      
    ]
\begin{verbatim}
You are a verification agent for table-based question answering.
You need to verify if the answer and reasoning for the given
question are correct based ONLY on the provided table data.

[Tables Used]
[Sampled Table Data (JSON Format)]

[Question-Answer Pair]
Question: [Generated Question Text]
Answer: [Generated Answer (JSON Format)]
Reasoning Steps: [Generated Reasoning Steps]
Question Type: [Designated Question Type]

Your task:
1. Check if the question is well-formed and genuinely requires multi-hop reasoning across 
MULTIPLE provided tables. Single-table questions are invalid.
2. Verify if the answer is accurate based only on the information present in the given tables. 
If the answer is incorrect, 'is_valid' must be 'false'.
3. Check if the 'tables_used' field correctly lists relevant tables and if at least
two tables were necessary.
4. Validate if the reasoning steps are logical, coherent, and correctly lead from the table 
data to the answer.

Respond with ONLY a valid JSON object (no introductory text, markdown formatting, 
or code blocks outside the JSON structure) containing the following keys:
{{
    "is_valid": true/false,
    "verification_comments": "Your detailed verification comments
                             explaining the validity/issues and
                             multi-table requirement.",
    "score": <an integer score from 0 to 10, where 10 is
              perfect adherence to all criteria>,
    "uses_multiple_tables": true/false
}}
\end{verbatim}
\end{tcolorbox}
\caption{LLM prompt for automated QA pair verification. Placeholders like `[Generated Question Text]` represent the actual data provided to the model.}
\label{fig:verification_prompt}
\end{figure}

\clearpage
\section{Visual Table Rendering Details}
\label{app:rendering_details}

As described in Section \ref{sec:rendering}, MTabVQA table images were generated with significant visual diversity to mimic real-world appearances. For each QA pair, the rendering process introduced controlled variations across several dimensions using 10 distinct styling themes, randomly selected per table. These themes systematically varied:

\begin{itemize}
    \item \textbf{Structure and Layout:}
    \begin{itemize}
        \item Column widths and row heights were adapted to content to ensure readability while introducing natural variations.
        \item The relative positioning of multiple table images within the final visual context presented to the model was also varied (e.g., tables rendered side-by-side, stacked vertically, or with other layout configurations).
    \end{itemize}
    \item \textbf{Appearance (Themes, Fonts, Styles):} The 10 distinct styling themes systematically manipulated the following:
        \begin{itemize}
            \item \textit{Color Schemes:} This included variations in header background colors (e.g., using specific hex codes like \texttt{\#4CAF50} (green), \texttt{\#1E88E5} (blue), \texttt{\#333} (dark grey)), cell background colors, text colors (e.g., white text on dark headers, black text on light backgrounds), and alternating row shading ('zebra striping' with colors like \texttt{\#f2f2f2}).
            \item \textit{Typography:} Different font families (e.g., common serif and sans-serif fonts) were used. Font weights were varied (e.g., bold headers, normal weight for cell content). Font sizes were adjusted within themes (e.g., a base size of \textbf{12pt} in one theme, with relative adjustments for headers).
            \item \textit{Styling Elements:} The presence, style, and color of grid lines were varied (e.g., solid lines, dashed lines, varying thickness, or minimalist themes with no grid lines). Cell padding was adjusted to control spacing within cells. Border styles for the overall table and individual cells were also diversified (e.g., \texttt{1px solid black}, \texttt{2px solid \#000}, or no borders).
        \end{itemize}
\end{itemize}
This deliberate introduction of visual diversity is key to challenging models on robust OCR and layout understanding across varied presentations before they engage in multi-tabular reasoning.

\clearpage
\section{Human Verification Interface}
\label{app:streamlit_interface}
\begin{samepage} 

Figure~\ref{fig:streamlit_interface} shows the interface of the Streamlit application used for the final human verification stage (Section~\ref{sec:verification}). This tool displayed the rendered table images, the generated question, the LLM-generated answer and reasoning, and the automated verification scores, allowing reviewers to make the final acceptance decision.

\begin{figure}[!htbp]
  \centering
  \vspace{1em} %
  
  \begin{subfigure}{\linewidth}
    \centering
    \includegraphics[width=0.85\linewidth]{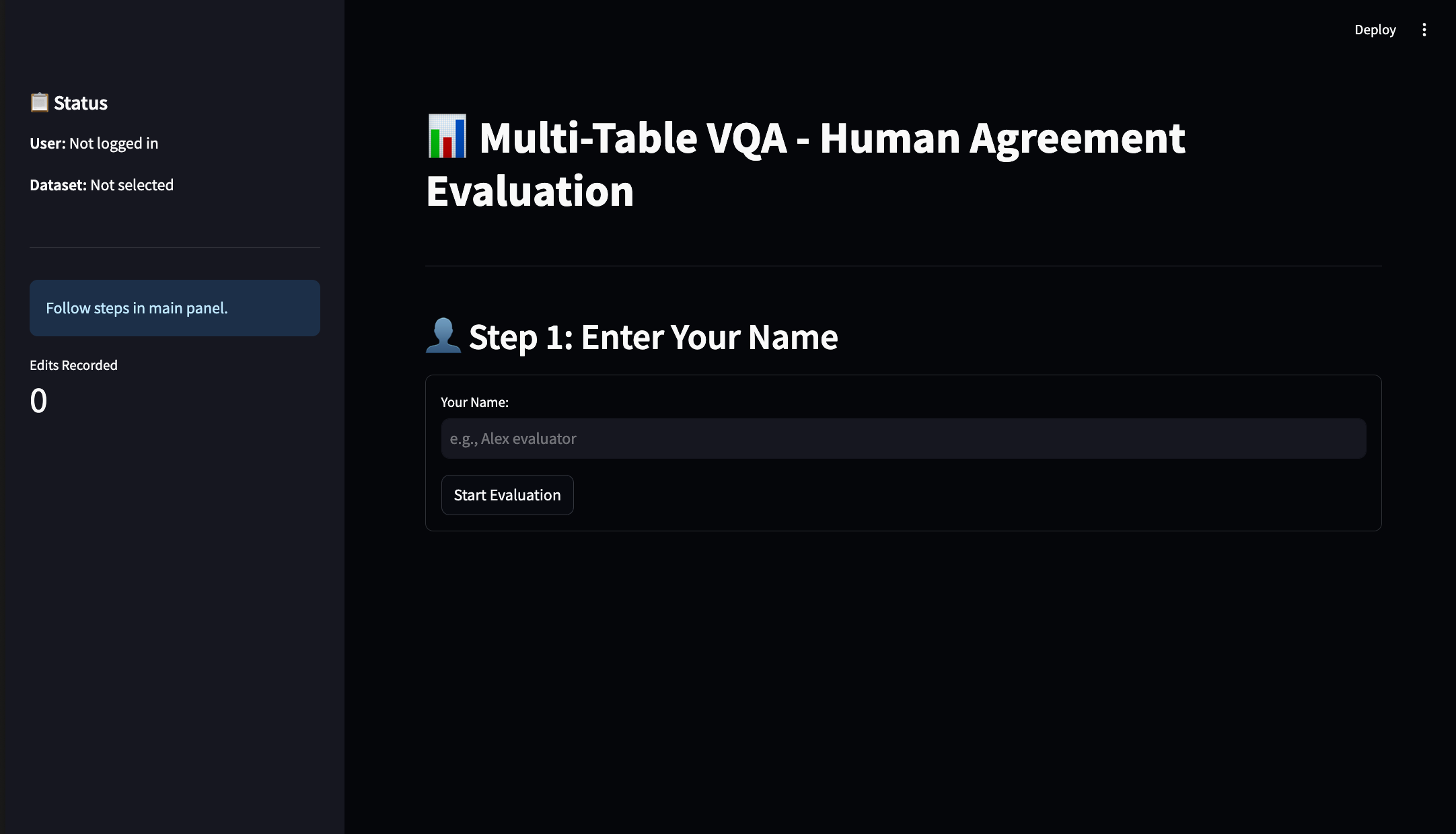}
    \caption{Initial login screen for evaluator identification.}
    \label{fig:streamlit_interface_login}
  \end{subfigure}
  
  \vspace{0.5em}
  
  \begin{subfigure}{\linewidth}
    \centering
    \includegraphics[width=0.85\linewidth]{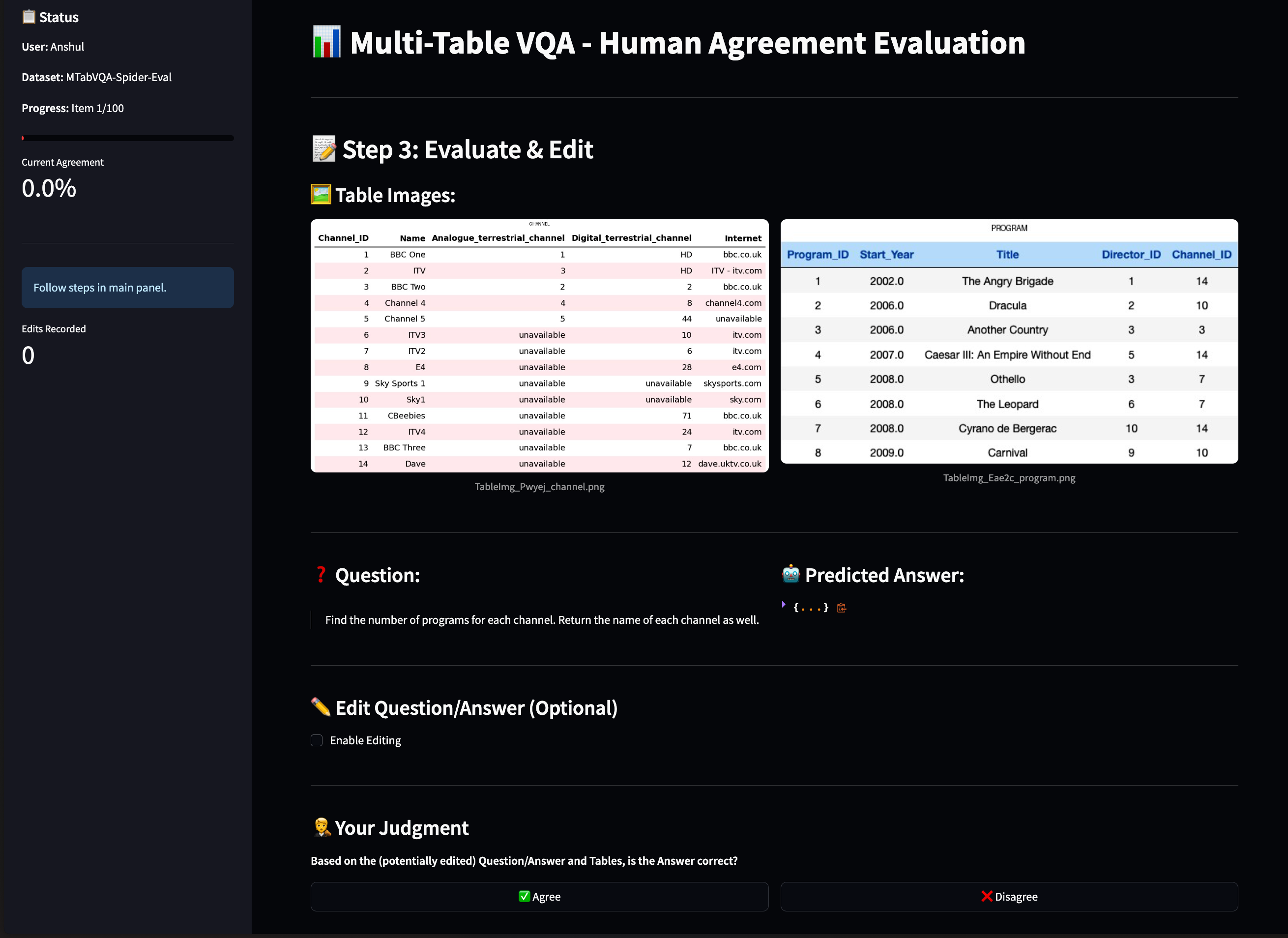}
    \caption{Main evaluation screen displaying table images, question, predicted answer, and reviewer judgment options.}
    \label{fig:streamlit_interface_eval}
  \end{subfigure}
  
  \caption{Screenshots of the Streamlit application interface used for human verification. Panel (\subref{fig:streamlit_interface_login}) shows the user login step, and panel (\subref{fig:streamlit_interface_eval}) presents the core evaluation interface with table images and QA details.}
  \label{fig:streamlit_interface}
\end{figure}
\end{samepage}
\clearpage

\section{GRPO Training Details}
\label{app:grpo_training}

This section provides additional details on the Group Relative Policy Optimization (GRPO) \cite{deepseekmath} experiments discussed in Section \ref{sec:post_training_analysis} for fine-tuning the Qwen2.5-VL-3B model. We utilized the EasyR1 framework\footnote{\url{https://github.com/hiyouga/EasyR1}} for these experiments, training for a total of 270 steps. The training was conducted on a subset of MTabVQA-Instruct derived from the Spider dataset (2,395 examples).
\begin{figure*}[h!]
  \centering
  \begin{subfigure}{0.32\linewidth}
    \includegraphics[width=\linewidth]{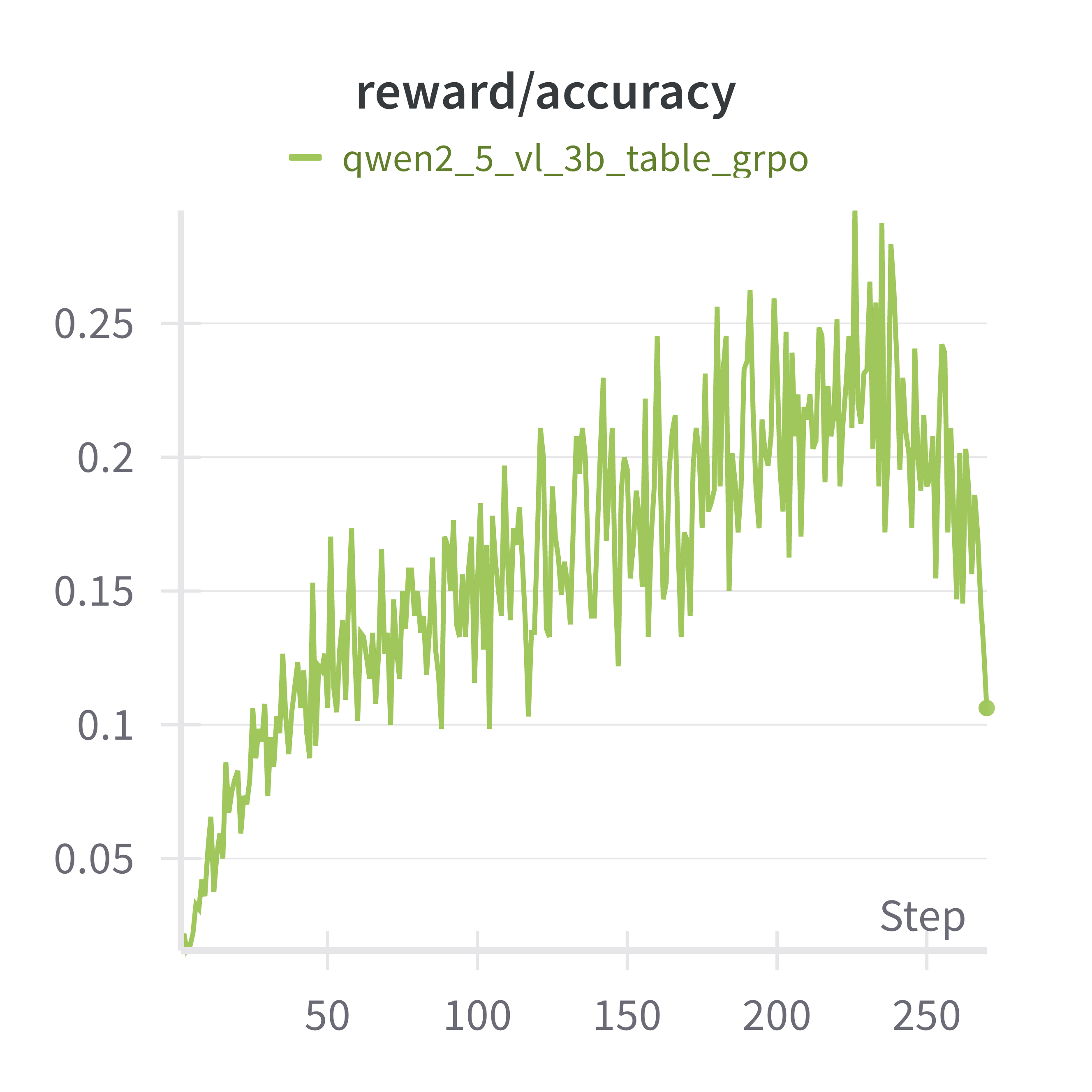} %
    \caption{Reward: Accuracy (EM)}
    \label{fig:grpo_reward_acc}
  \end{subfigure}
  \hfill
  \begin{subfigure}{0.32\linewidth}
    \includegraphics[width=\linewidth]{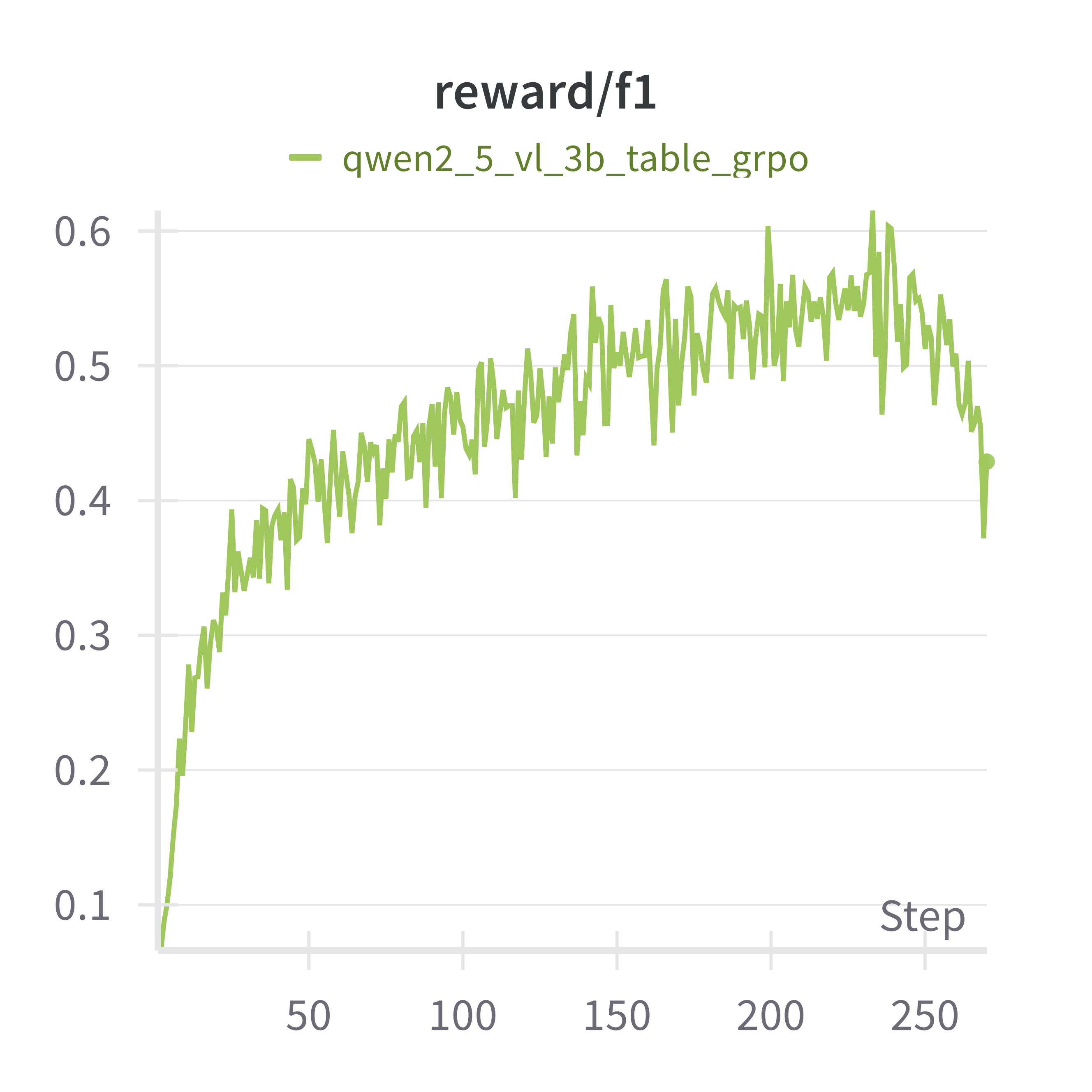}
    \caption{Reward: F1 Score}
    \label{fig:grpo_reward_f1}
  \end{subfigure}
  \hfill
  \begin{subfigure}{0.32\linewidth}
    \includegraphics[width=\linewidth]{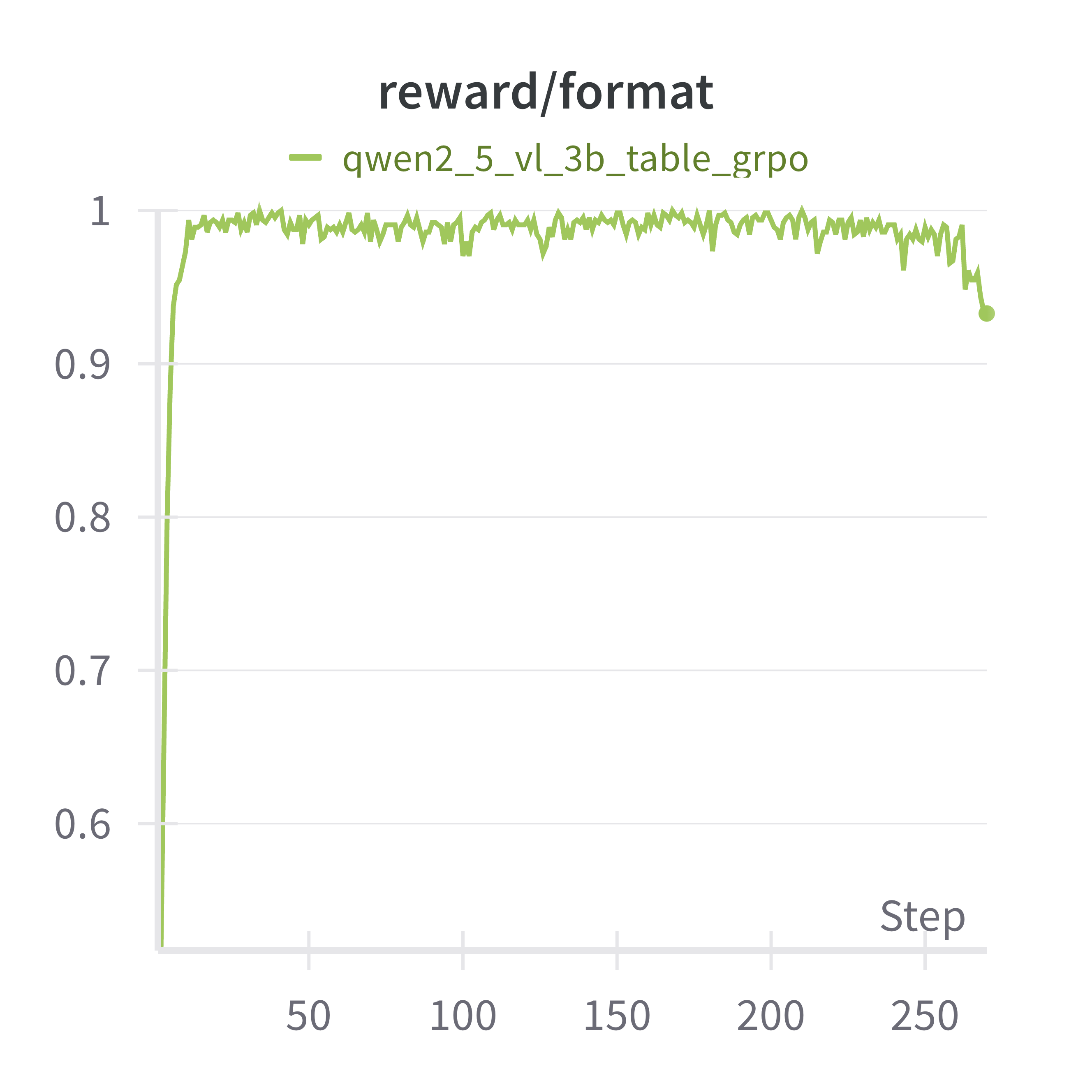} %
    \caption{Reward: Structural Format}
    \label{fig:grpo_reward_struct_format}
  \end{subfigure}

  \vspace{1em} %

  \begin{subfigure}{0.32\linewidth}
    \includegraphics[width=\linewidth]{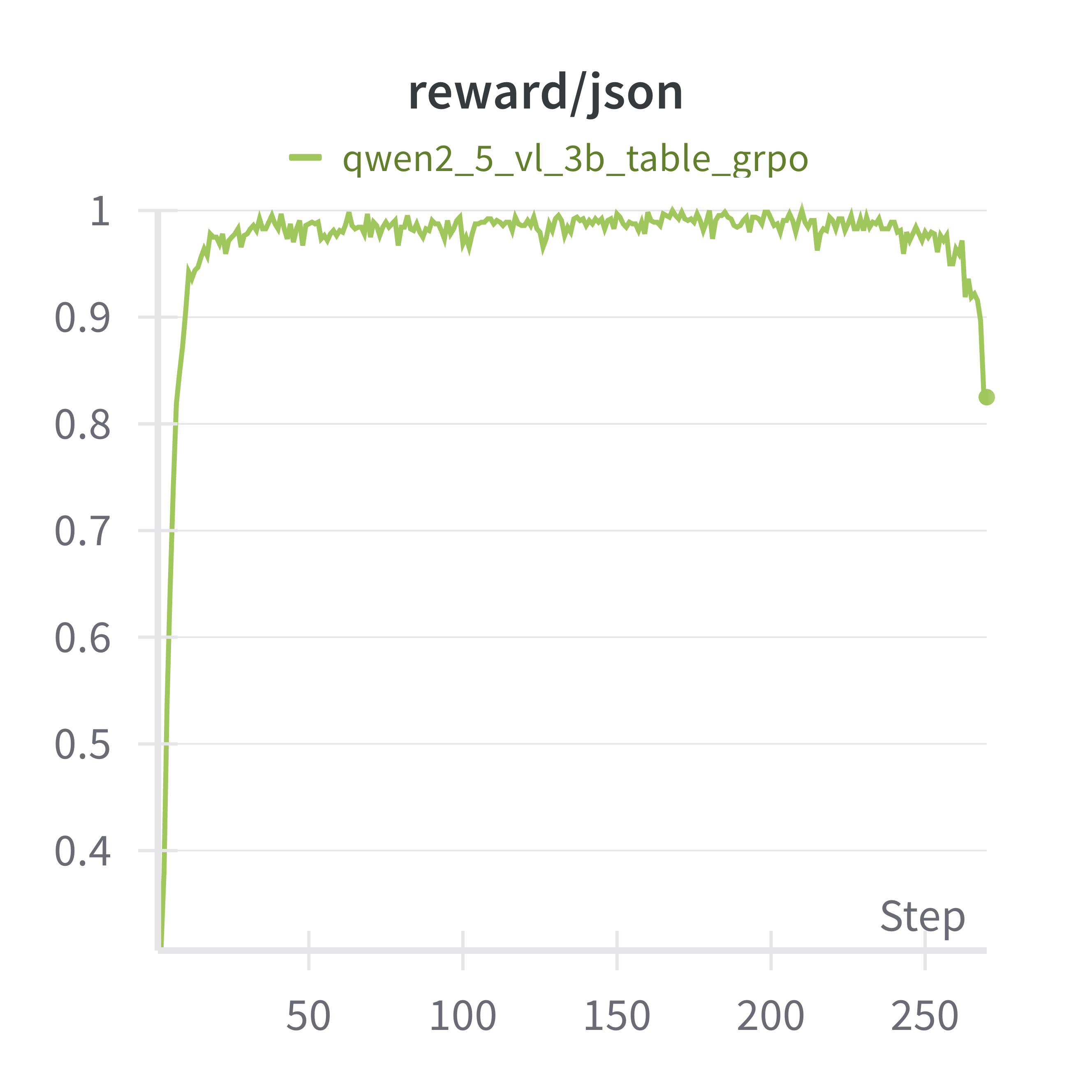} %
    \caption{Reward: JSON Format}
    \label{fig:grpo_reward_json_format}
  \end{subfigure}
  \hfill
  \begin{subfigure}{0.32\linewidth}
    \includegraphics[width=\linewidth]{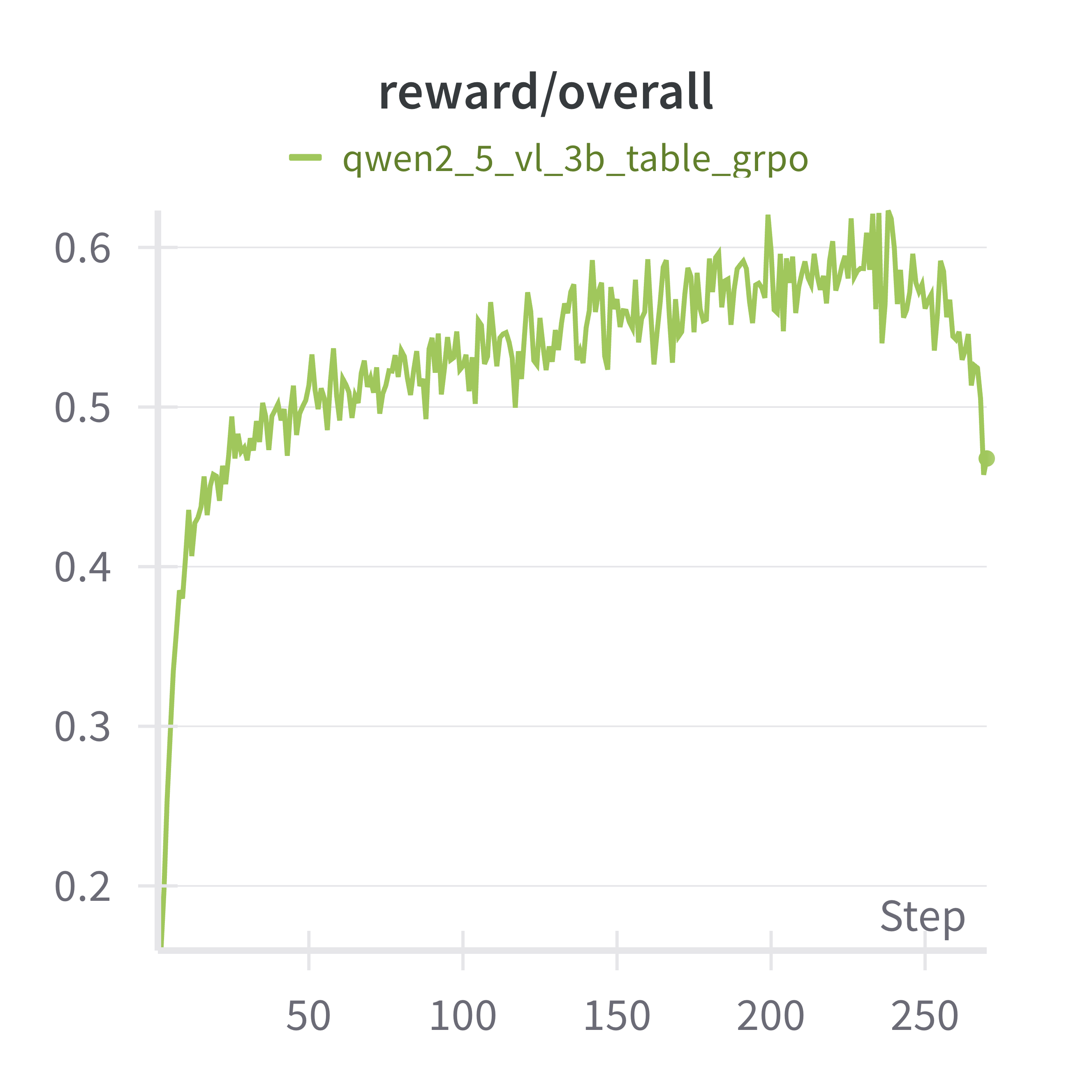} %
    \caption{Reward: Overall}
    \label{fig:grpo_reward_overall_combined}
  \end{subfigure}
  \caption{GRPO training reward component curves for Qwen2.5-VL-3B over 270 training steps. These plots illustrate the learning progress for content accuracy (EM, F1), structural format adherence, JSON validity, and the combined overall reward.}
  \label{fig:grpo_reward_plots}
\end{figure*}

\textbf{Reward Function:} The reward function for GRPO was designed to encourage both semantic correctness and proper output formatting. It was a composite score derived from:
\begin{itemize}
    \item \textbf{Content Correctness:} Assessed by the weighted sum of Exact Match (EM) and F1 score between the generated answer and the ground truth.
    \item \textbf{Format Adherence:} This included two components:
        \begin{itemize}
            \item \textit{Structural Format Score:} A binary score indicating whether the model's output correctly included the required `<think>` and `<answer>` tags.
            \item \textit{JSON Format Score:} A binary score indicating whether the content within the `<answer>` tags was valid JSON.
        \end{itemize}
\end{itemize}
The overall reward signal aimed to maximize these components, guiding the model towards generating accurate and well-formatted responses.

Figure \ref{fig:grpo_reward_plots} shows the progression of various reward components during the GRPO training process. The plots for `reward/accuracy` (EM) and `reward/f1` show a general upward trend, indicating learning of content correctness. The `reward/format` and `reward/json` plots demonstrate that the model quickly learned to adhere to the specified output structure. The `reward/overall` plot reflects the combined learning signal. The final checkpoint used for evaluation was selected based on the highest `reward/overall` achieved during training. These settings were chosen to balance training stability, computational efficiency, and exploration during the reinforcement learning process for the multi-tabular visual question answering task, aiming for both accurate content and correctly formatted output. Key GRPO training parameters are summarized in Table \ref{tab:grpo_hyperparams_concise}.

\begin{table}[htbp]
\centering
\begin{tabular}{@{}ll@{}}
\toprule
\textbf{Parameter} & \textbf{Value} \\
\midrule
\multicolumn{2}{l}{\textbf{Core Algorithm}} \\
\quad Advantage Estimator & GRPO \\
\quad KL Coefficient ($\lambda_{KL}$) & 0.01 \\
\midrule
\multicolumn{2}{l}{\textbf{Training Setup}} \\
\quad Base Model & Qwen/Qwen2.5-VL-3B-Instruct \\
\quad Training Data & MTabVQA-Instruct (Spider Subset) (2,395 ex.) \\ %
\quad Max Training Steps & 270 \\
\quad Total Epochs & 15 \\
\quad Rollout Batch Size & 128 \\
\midrule
\multicolumn{2}{l}{\textbf{Actor Model (Qwen2.5-VL-3B)}} \\
\quad Learning Rate & 1e-06 \\
\quad Optimizer & AdamW (BF16) \\
\quad Global Update Batch Size & 32 \\
\midrule
\multicolumn{2}{l}{\textbf{Rollout Generation}} \\
\quad Temperature (Training) & 1.0 \\
\quad Top-p (Training) & 0.99 \\
\quad Num. Generations per Prompt (n) & 5 \\
\midrule
\end{tabular}
\caption{GRPO Hyperparameters for Qwen2.5-VL-3B Fine-tuning.}
\label{tab:grpo_hyperparams_concise} %
\end{table}

\clearpage
\section{Model Evaluation and Generation Prompts}
\label{app:evaluation_prompts}

This section details the system prompts used for evaluating and generating responses from the Vision-Language Models (VLMs) in different experimental settings.

\subsection{Standard Zero-Shot Evaluation Prompt}
\label{app:standard_eval_prompt}

For standard zero-shot evaluations of VLMs (Section \ref{sec:benchmarking}), including proprietary models and open-source baselines before specific post-training, the following system prompt was used. This prompt instructs the model on how to interpret multi-tabular image data, reason about the question, and provide an answer strictly in the specified JSON format.

\begin{tcolorbox}[
    title={System Prompt: Zero-Shot Evaluation},
    width=\linewidth, %
    colback=gray!5!white,
    colframe=black!60!black,
    boxrule=0.5pt,
    arc=1mm,
    fontupper=\small,
    ]
\begin{verbatim}
You are an intelligent assistant capable of understanding and reasoning about multi-tabular data 
given as images, each table is one image. You will be presented with one or more tables containing 
information on a specific topic. 
You will then be asked a question that requires you to analyze the data in the table(s) and provide a 
correct answer in strict required format.

Your task is to:

1. Carefully examine the provided table(s) Pay close attention to the column headers, the data types 
within each column, and the relationships between tables if multiple tables are given.
2. Understand the question being asked. Identify the specific information being requested and 
determine which table(s) and columns are relevant to answering the question.
3. Extract the necessary information from the table(s). Perform any required filtering, joining, 
aggregation, or calculations on the data to arrive at the answer.
4. Formulate a clear and concise answer in natural language. The answer should be directly responsive 
to the question and presented in a human-readable format. It may involve listing data, presenting 
a single value, or explaining a derived insight.
5. Do not include any SQL queries in the answer. But you can use it internally, to come up with answer.
6. Be accurate and avoid hallucinations. Your answer should be completely based on the data 
in the provided table(s). Do not introduce any external information or make assumptions not supported
by the data.
7. Be specific and follow the instructions in the question. If the question ask to get specific 
columns, return only mentioned columns.
8. If the question is unanswerable based on the provided tables, state "The question cannot be 
answered based on the provided data.
9. Please provide only the answer which has been asked, without any additional text (try to use 
few tokens). However, take the time to think and reason before
giving your answer. Also, try to provide an answer even if you are unsure.
10. Provide the answer in JSON format with given response schema as given 
[['ans1','ans2'],['ans3','ans4']]. Respond only with valid JSON format.

Take your time to understand the question. Break it down into smaller steps. Come up with 
an answer and examine your reasoning. Finally, verify your answer. 
you need to extract answers based on the given multi-hop question [Question] and given multiple tables
[TABLE1], and [TABLE2]. Please only output the results without any other words.
Return the answer in the following JSON format.

Return the answer in JSON schema: : {
              "type": "json_schema",
              "json_schema": {
                  "name": "Response",
                  "type": "object",
                  "properties": {
                      "data": {
                          "type": "array",
                          "items": {"type": "array", "items": {"type": "string"}},
                      }
                  },
                  "required": ["data"],
                  "additionalProperties": False,
              },
          }
\end{verbatim}
\end{tcolorbox}
\vspace{1em} %

\subsection{Chain-of-Thought (CoT) Evaluation Prompt}
\label{app:cot_eval_prompt}

For the Chain-of-Thought (CoT) prompting experiments (Section \ref{sec:post_training_analysis}), a modified system prompt was used. This prompt explicitly instructs the model to first generate a step-by-step reasoning process (the chain of thought) and then provide the final answer.

\begin{tcolorbox}[
    title={System Prompt: Chain-of-Thought (CoT)},
    width=\linewidth, %
    colback=gray!5!white,
    colframe=black!60!black,
    boxrule=0.5pt,
    arc=1mm,
    fontupper=\small,
    ]
\begin{verbatim}
You are an intelligent assistant capable of understanding and reasoning about multi-tabular data
given as images, each table potentially being one image. You will be presented with one or more tables
containing information on a specific topic. You will then be asked a question that  requires you to 
analyze the data in the table(s) and provide a correct answer in the strictly required format.

Your task is to:

1.  Carefully examine the provided table(s): Pay close attention to the column headers, the 
data types within each column, and the relationships between  tables if multiple tables are given.
2.  Understand the question being asked: Identify the specific information being requested and 
determine which table(s) and columns are relevant to answering the question.
3.  Reason step-by-step (Chain of Thought): Before generating the final answer,
formulate a clear chain of thought outlining how you identified the relevant data, 
performed necessary operations (filtering, joining, aggregation, calculations), 
and arrived at the result. This reasoning is crucial and MUST be included in the final output.
4.  Extract the necessary information from the table(s): Perform any required filtering, joining, 
aggregation, or calculations on the data based on your chain of thought to arrive at the answer.
5.  Do not include any SQL queries in the final answer JSON. You can use SQL logic internally during 
your reasoning (Chain of Thought), but the final output should not contain raw SQL code.
6.  Be accurate and avoid hallucinations: Your answer must be completely based on the data in
the provided table(s).
. Provide the output strictly in the specified JSON format: The output must be a single JSON object 
containing two keys: `chain_of_thought` (a string detailing your reasoning steps) and `data` 
(an array of arrays containing the answer).

Your entire response must be ONLY a valid JSON string conforming to the schema below.
JSON Schema:
```json
{
  "type": "object",
  "properties": {
    "chain_of_thought": {
      "type": "string",
      "description": "A detailed step-by-step explanation of the reasoning process 
      used to arrive at the answer."
    },
    "data": {
      "type": "array",
      "items": {
        "type": "array",
        "items": {
          "type": "string" 
        }
      },
      "description": "The result data, formatted as an array of arrays, where each 
      inner array represents a row."
    }
  },
  "required": [
    "chain_of_thought",
    "data"
  ],
  "additionalProperties": False
}
```

Take your time to understand the question and the data. Break the problem down using Chain of Thought. 
Construct the final JSON containing both your reasoning and the extracted data. Verify 
your answer and the format before outputting. Remember to output ONLY the JSON string.
\end{verbatim}
\end{tcolorbox}
\vspace{1em}

\subsection{GRPO Thinking Prompt }
\label{app:grpo_prompt}

For the Group Relative Policy Optimization (GRPO) training and generation (Section \ref{sec:post_training_analysis} and Appendix \ref{app:grpo_training}), the prompt was used. This prompt is similar to the CoT prompt in that it requires an internal reasoning process (`<think>...</think>`) before the final answer, but it is specifically tailored for the GRPO framework, which often involves distinct markers for thought processes versus final outputs used in reward calculation. The final answer is expected within `<answer>...</answer>` tags in a specific JSON format.

\begin{tcolorbox}[
    title={System Prompt: GRPO Thinking Prompt },
    width=\linewidth, %
    colback=gray!5!white,
    colframe=black!60!black,
    boxrule=0.5pt,
    arc=1mm,
    fontupper=\small,
    ]
\begin{verbatim}
You are an intelligent assistant capable of understanding and reasoning about multi-tabular data 
given as images, each table is one image. You will be presented with one or more tables containing 
information on a specific topic. 
You will then be asked a question that requires you to analyze the data in the table(s) and 
provide a correct answer in strict required format using multi-hop reasoning.

Your task is to:

1. Carefully examine the provided table(s) Pay close attention to the column headers, the data types 
within each column, and the relationships between tables if multiple tables are given.
2. Understand the question being asked. Identify the specific information being requested and 
determine which table(s) and columns are relevant to answering the question.
3. Extract the necessary information from the table(s). Perform any required filtering, joining,
aggregation, or calculations on the data to arrive at the answer.
4. Formulate a clear and concise answer in natural language. The answer should be directly 
responsive to the question and presented in a human-readable format. 
It may involve listing data, presenting a single value, or explaining a derived insight.
5. Do not include any SQL queries in the answer. But you can use it internally, to come up with answer.
6. Be accurate and avoid hallucinations. Your answer should be completely based on the data in the
provided table(s). Do not introduce any external information or make assumptions not 
supported by the data.
7. Provide the answer in JSON format with given response schema as given
[['ans1','ans2'],['ans3','ans4']].Respond only with valid JSON format, as shown in the example above.

Strictly, Give answer in this format, using the example below as reference: 

You FIRST think about the reasoning process as an internal monologue and then provide the final answer.
The reasoning process MUST BE enclosed within <think> </think> tags. You will be presented with one
or more tables containing information on a specific topic.You will then be asked a question that 
requires you to  analyze the data in the table(s) and provide a correct answer.The final answer MUST BE
put in <answer> </answer> in json format.
Example JSON format inside  <answer>{"data": [[‘ans1’, ‘ans2’], [‘ans3’, ‘ans4’]]}</answer>.
\end{verbatim}
\end{tcolorbox}

\end{document}